\documentclass[conference]{IEEEtran}
\IEEEoverridecommandlockouts
\usepackage{cite}
\usepackage{amsmath,amssymb,amsfonts}
\usepackage{algorithmic}
\usepackage{graphicx}
\usepackage{textcomp}
\usepackage{xcolor}
\usepackage{url}
\usepackage{listings}
\usepackage{lipsum}
\usepackage{orcidlink}
\usepackage[T1]{fontenc}
\usepackage[inline]{enumitem}
\usepackage{siunitx}
\usepackage{tikz}
\usepackage{pifont}
\newcommand{\redx}{\textcolor{red}{\ding{55}}}
\definecolor{darkgreen}{RGB}{0,128,0}
\newcommand{\greencheck}{\textcolor{darkgreen}{\checkmark}}
\usepackage{booktabs}
\usepackage{glossaries}
\newacronym{API}{API}{application programming interface}
\newacronym{IMU}{IMU}{inertial measurement unit}

\def\BibTeX{{\rm B\kern-.05em{\sc i\kern-.025em b}\kern-.08em
    T\kern-.1667em\lower.7ex\hbox{E}\kern-.125emX}}
\begin{document}

\title{Lowering Barriers to Entry for Fully-Integrated Custom Payloads on a DJI Matrice
}

\author{\IEEEauthorblockN{Joshua Springer}
\IEEEauthorblockA{\textit{Department of Computer Science} \\
\textit{Reykjavik University}\\
Reykjavik, Iceland \\
\orcidlink{0000-0003-0137-1770} \href{https://orcid.org/0000-0003-0137-1770}{orcid.org/0000-0003-0137-1770}
}
\and
\IEEEauthorblockN{Gylfi Þór Guðmundsson}
\IEEEauthorblockA{\textit{Department of Computer Science} \\
\textit{Reykjavik University}\\
Reykjavik, Iceland \\
\orcidlink{0000-0003-0846-6617} \href{https://orcid.org/0000-0003-0846-6617}{orcid.org/0000-0003-0846-6617}}
\and
\IEEEauthorblockN{Marcel Kyas}
\IEEEauthorblockA{\textit{Department of Computer Science} \\
\textit{Reykjavik University}\\
Reykjavik, Iceland \\
\orcidlink{0000-0003-1018-3413} \href{https://orcid.org/0000-0003-1018-3413}{orcid.org/0000-0003-1018-3413}}
}

\maketitle

\begin{abstract}
    Consumer-grade drones have become effective multimedia collection tools,
spring-boarded by rapid development in embedded CPUs, GPUs, and cameras.
They are best known for their ability to cheaply collect
high-quality aerial video,
3D terrain scans,
infrared imagery, etc.,
with respect to manned aircraft.
However, users can also create and attach custom sensors, actuators, or computers,
so the drone can collect different data,
generate composite data,
or interact intelligently with its environment,
e.g., autonomously changing behavior to land in a safe way,
or choosing further data collection sites.
Unfortunately, developing custom payloads is prohibitively difficult for many researchers
outside of engineering.
We provide guidelines for how to create a sophisticated computational payload that integrates
a Raspberry Pi 5 into a DJI Matrice 350.
The payload fits into the Matrice's case like a typical DJI payload (but is much cheaper),
is easy to build and expand (3D-printed),
uses the drone's power and telemetry,
can control the drone and its other payloads,
can access the drone's sensors and camera feeds,
and can process video and stream it to the operator via the controller in real time.
We describe the difficulties and proprietary quirks we encountered,
how we worked through them,
and provide setup scripts and a known-working configuration for others to use.



\end{abstract}


\section{Introduction}
The rise of the modern smartphone caused a rapid expansion of large image and video collections in two decades.
This, in turn, spurred the development of large-scale image and video processing algorithms.
Cheap camera drones take this technology into the air to places people cannot access, thereby diversifying these collections.
Consumers, journalists, and cinematographers record video from the air easily.
Emergency responders use such drones for surveying and rescue tasks.

Being able to equip the drone with a custom sensor and or video processing capability
(i.e., creating a {\it custom payload}) is essential for many data collection tasks. 
At first glance, it may seem desirable to build your own (DIY) custom drone.
The hardware is cheap and readily available,
and open source flight control systems~\cite{ardupilot,px4}
exist that can serve as the basis for your platform development.
With full access to underlying code, implementing a custom payload should be easy. 
However, without a team of supporting experts,
the financial savings will quickly evaporate due to the time investment,
as undertaking such a DIY project is no trivial task~\cite{marked_landing_pads_lava_flows}.
Building a custom drone requires a working knowledge of control systems, electronics,
telemetry, manufacturing/assembly, safe testing practices etc.,
before a developer can even broach the topic of developing an application-specific payload~\cite{cascade_wood, volcanology_wood}.
In cases where we do not focus on drone construction or low-level control, but instead on data collection and/or processing, the DIY route requires us to contend with the full complexity and practical overhead of the drone platform.
Thus, for many, the DIY overhead is prohibitive.

Alternatively, commercial drones provide a stable and reliable sensor platform out of the box,
and the skill requirements for reliable operation are minimal,
making them ideal for small teams or researchers in unrelated domains.
However, they are much more expensive, and their computational systems are closed source, making custom payload development difficult and slow.
The main challenge is integrating the custom payload into the commercial
drone platform,
which is neither straightforward nor a high priority for the drone manufacturer,
as it is neither a particularly common nor profitable aspect of their business.
The result is that the software \gls{API} is essentially a black-box
with only limited functionality and documentation exposed to the public.
This can be a prohibitively difficult barrier to entry for researchers without
specializations in engineering or computer science.
At the extremes, this can mean they prefer to use the drone only to physically
carry a custom payload, and avoid using the drone's computational system
altogether, even if it means adding redundant power, sensor and/or other
equipment, because the equipment onboard the drone is too difficult to access.

We share our experience in developing a {\it custom payload} for DJI Matrice 350 and demonstrate its operation in the field. 
It is our hope that our work may serve as guidelines for others to get the most out of their commercial platform. 
The payload is a Raspberry Pi 5 whose functionality is to add autonomous precision landing capabilities to our drone. This requires the following functionalities and sensor data from the drone:
\begin{enumerate*}[label={\arabic*)}]
\item \SI{5}{\volt} DC power to run Pi;
\item real-time access to the drone's camera feeds (including stereovision depth cameras)
and flight controller data (e.g., \gls{IMU}, altimeter, etc.);
\item control of the drone's movement and of its other payloads; 
\item the orientation of its gimbal-mounted cameras;
\item the ability to process the drone's video in real-time; and 
\item transmit and display it on the operator's hand-held controller.
\end{enumerate*}
We explain how we implement these functionalities on 
our Matrice 350 and describe the difficulties we overcame despite the limited documentation and tutorials and the proprietary quirks of the available \gls{API}, and we show the performance of the system in an in-flight scenario.

\section{Background}
\subsection{DJI Matrice 350}

The Matrice 350 is DJI's flagship enterprise drone that is ideal for remote sensing applications because of its portability, long flight time compared to other commercial drones (about 55 minutes), water-resistance such that it can fly in light rain, long distance telemetry range, payload capacity of \SI{2.7}{\kilo\gram},
and availability of proprietary payloads such as wide-angle, zoom, and thermal cameras, LIDAR, laser rangefinders, and the ability to add custom payloads~\cite{m350}.
Custom payloads can access the drone's power and telemetry systems, video streams
from onboard or payload cameras, and many data topics describing the drone's
conditions.
Existing custom payloads include gas samplers, water samplers, 5G connectors, and emergency parachutes~\cite{psdk_intro}.

\subsection{DJI SDKs}

DJI provides 3 SDKs: Mobile SDK (MSDK)~\cite{msdk_intro}, Onboard SDK (OSDK)~\cite{osdk_github},
and Payload SDK (PSDK)~\cite{psdk_intro}.
The MSDK allows users to program Android or iOS applications to run their own code on the drone's controller or connected tablet/phone.
The applications can view the drone's video and data feeds, and can control the drone.
The OSDK and PSDK offer the ability to embed the processing onto the drone, such that the tasks do not depend on the controller's data link,
which makes everything faster and more reliable.
The PSDK is newer and is what we use here, but both the OSDK and PSDK are set up as cmake projects in C and C++ and target a Manifold 2 (a DJI onboard computer),
and the NVIDIA Jetson series, although tutorials have been provided for the Raspberry
Pi 4.
DJI provides several code samples outlining common use cases for the PSDK.
However, there is still a high barrier to entry in using the SDKs,
such that they can be prohibitively difficult to use for people outside of the domain
of software development.
Moreover, many nuances of the PSDK are not directly addressed in the documentation,
such that some essential information can only be found through communication with DJI
support.

\section{Methods}
Our goal is to make a generalized payload that can be easily adapted to the requirements of
researchers from various domains,
to lower the (currently quite high) barrier to entry for using custom payloads on DJI drones.
We do this from the point of view of our specific application, which is intended to use the drone's
onboard sensors to identify safe landing sites in lava fields, execute the landings autonomously,
and ensure the human operator is informed about the system's conditions and intentions in real time
in case some intervention is necessary.
This involves the creation/use of some fundamental infrastructure that generalizes to many custom
payloads that might be desired by other researchers,
as specified below:

\begin{enumerate}[label=\textbf{Req. \arabic*:}, ref=\arabic*,leftmargin=*]
    \item \textbf{Portability.} The payload quickly mounts onto the drone without assembly and can be carried in the drone's case as a typical DJI payload.
    \label{requirement:dji_style_portability}
    \item \textbf{Ease of software development.} The payload uses common computational hardware.
    \label{requirement:accessible_hardware}
    \item \textbf{Power/telemetry integration.} The payload uses the drone's power and telemetry systems,
    thereby reducing the amount of redundant equipment. \label{requirement:power_telemetry}
    \item \textbf{Flight/payload control.} The payload can control the behavior of the drone and of other payloads, e.g., it can aim other payload cameras.
    \label{requirement:control_drone_and_payloads}
    \item \textbf{Access to drone's sensors.} The payload can get input from the drone's many onboard sensors and from other payloads, e.g., composite pose estimates, GPS, altitude over ground, distance/obstacle sensors, etc. \label{requirement:onboard_sensors}
    \item \textbf{Access to camera feeds.} The payload can view RGB/stereo video from the drone and other payloads.\label{requirement:onboard_video_feeds}
    \item \textbf{Real-time video processing/streaming.} The payload can process sensor video and can relay this video to the pilot's handheld controller in real time.\label{requirement:process_transmit_video}
\end{enumerate}

\subsection{Physical Setup}
For \textbf{Req.~\ref{requirement:dji_style_portability}}, 3D printing is the easiest method available to us for developing a small form-factor
case for these components that is adequately weatherproof.
Physical space for safe transportation in the drone's case limits the case to a 
bounding box of about 15 cm x 15 cm x 11 cm.
We print a simple case in the shape of a tapered dodecagonal prism with air holes at
the bottom, and each component mounts vertically to take efficient advantage of the
space.
The components can be removed independently using a single screw and pulling vertically upwards.
The payload attaches to the drone with 3D-printed quick-release latches.
The payload case is designed in OpenSCAD and the files are available on GitHub~\cite{matrice_parts_repo}.

To address \textbf{Req.~\ref{requirement:accessible_hardware}}, we use the Raspberry Pi 5~\cite{rpi5}
because it has support for a lot of software and sensors,
is inexpensive, and relatively easy to use given its developed community.
The mount point for onboard computers and many other custom sensors is on top of the
Matrice, and has access to 2 payload ports:
an E-port (for onboard computers)
and a SkyPort V2 (for typical payloads)
-- both of which use the PSDK but have different capabilities~\cite{psdk_function_set},
as shown in Table~\ref{table:psdk_functionalities}.
\begin{table}[hb]
\centering
\begin{tabular}{llllll}
\toprule
\textbf{Port} & \textbf{Req 3.} & \textbf{Req. 4} & \textbf{Req. 5} & \textbf{Req. 6} & \textbf{Req. 7} \\
\midrule
E-port        & \greencheck & \greencheck & \greencheck & \greencheck & \redx \\
SkyPort V2    & \greencheck & \redx & \greencheck & \redx & \greencheck \\
\bottomrule \\
\end{tabular}
\caption{Per-port PSDK functionalities.}
\label{table:psdk_functionalities}
\end{table}
The E-port provides flight and payload control, the ability to integrate into the drone's
power and telemetry systems, access to the drone's instruments (including video streams),
but does not offer the ability to stream video to the controller for the pilot to view
in real time.
On the other hand, the SkyPort can stream video to the controller but lacks the ability to
access video feeds from the drone and other payloads, and also lacks the ability to control
the drone and other payloads.
Therefore, our payload requires accessing both ports.
Each PSDK application can only connect to one port at a time,
so the payload must run two separate applications in parallel.
The Raspberry Pi 5 is the central component.
Each application requires two data connections to the drone -- a UART for low-bandwidth
communication and configuration, and either network or USB bulk connection for high-bandwidth
data transfer such as real-time video.
The E-port adapter board provides power to the Pi both over USB C and via a DC-DC
converter that converts 12V to 5V, fulfilling \textbf{Req.~\ref{requirement:power_telemetry}}.
It also provides a UART interface to the Pi and communicates with the Pi over the
USB-C line.
The UART on the SkyPort adapter board connects to the Pi via a USB-to-UART converter
and also via an Ethernet cable.


\subsection{E-port Application: Retrieving Camera Feeds}
\label{section:eport_application}

We cover the main points for setting up an application that can access
camera and sensor feeds via the E-port.
We further provide a setup script \texttt{configure.sh} to automate the
procedure, which contains all relevant details~\cite{configure_sh}.
This script carries out all necessary configuration described below in order to create the
E-port application and has been tested with V3.8.1 of the PSDK on the 32-bit Raspbian OS image from 15 March 2024.
The remaining specifications can be found in our GitHub fork~\cite{psdk_fork} of DJI's PSDK repository~\cite{psdk}.

We use DJI's E-port development board, which provides a UART interface for configuration
and low-bandwidth communication, and a USB C port for high-bandwidth communication,
i.e., retrieving video from the drone.
We connect both interfaces to the Raspberry Pi -- UART to the GPIO UART, and USB C to the 
Raspberry Pi's USB C port.
The only software configuration required for the UART is to set a macro within the PSDK that
specifies which UART to use.
With this configured, the PSDK application can connect to the drone,
control the drone's flight and interact with other payloads
(\textbf{Req.~\ref{requirement:control_drone_and_payloads}}),
and
subscribe to data topics
(\textbf{Req.~\ref{requirement:onboard_sensors}}).

To retrieve video feeds (\textbf{Req.~\ref{requirement:onboard_video_feeds}}),
we set up two USB bulk virtual devices on the Raspberry Pi.
These allow the drone to write files directly to the Pi using the FunctionFS protocol over the
Pi's USB C port.
We take inspiration from a Chinese language tutorial from DJI on how to set up USB bulk
devices for the Raspberry Pi 4B on the Matrice 30~\cite{dji_tutorial_rpi_m30}.
This is the closest documentation we could find from DJI for implementing our system,
but it still differed in the drone model, and Raspberry Pi model,
which are major factors requiring significant consideration and a unique configuration.
The script declares the Pi as a multifunction composite USB gadget with RNDIS and USB Bulk
functionality and sets specific vendor and product IDs such that the drone can identify it
unambiguously.
It creates a directory for a FunctionFS filesystem and calls a custom program to manage it,
which subsequently creates endpoints for management/configuration, input, and output.
The PSDK allows users to set 2 macros for FunctionFS file systems, and it appears that the
Matrice 350 uses them both
(one for RGB video and one for stereo video),
while the tutorial for the M30 only provides one.
Therefore, we edit the content of the tutorial to create these 2 file systems instead of 1.
Finally, we configure the PSDK application explicitly to use both a UART and USB bulk device,
add our application credentials, and compile the PSDK code.
For this demonstration, we simply run a sample application provided by DJI that implements
API calls for viewing RGB/stereo video, for subscribing to data topics, and for controlling the
drone and other payloads.

\subsection{SkyPort Application: Streaming Video to Controller}

In this section, we cover the main points for setting up an application that streams the desktop of
the Raspberry Pi, such that any visualization can simply be shown in a typical on-screen window
and will appear on the controller.
This strategy allows for the video streaming application to run completely independently of the E-port
application and for easily changing the video source that is streamed.

We connect the application to the drone via a SkyPort development board that offers a UART and
Ethernet port.
The UART functions as a low-bandwidth data connection,
and the Ethernet port provides the high-bandwidth connection for video,
as on the E-port dev board.
We configure the application to use both a UART and network device,
and we set the UART to be the device address of the USB-to-UART converter,
and the network connection to be the main Ethernet interface.
The payload does not take power from this port because it is already powered by the E-port.

We fulfill \textbf{Req.~\ref{requirement:process_transmit_video}} using the camera emulation sample given by the PSDK to stream video to the controller.
The first step is to switch from the Wayland desktop environment to the X11 environment because it
allows for easier screengrabbing via \texttt{ffmpeg}.
In the PSDK application for the SkyPort V2, we create an \texttt{ffmpeg} subprocess that 
provides a video stream of the desktop as raw frames.
In software, we encode the frames to \texttt{.h264} format and forward them to the drone
using the \texttt{DjiPayloadCamera\_SendVideoStream} function, chopping each frame into multiple
packets if it is too large.
It then appears as a video source on the drone's controller.
For a small level of interaction with the Pi itself, we also implement mouse clicks using
\texttt{xdotool} when the pilot clicks on a particular part of the controller's screen.

\section{Results}
\subsection{Requirements}

\begin{figure}
    \centering
    \includegraphics[height=4cm]{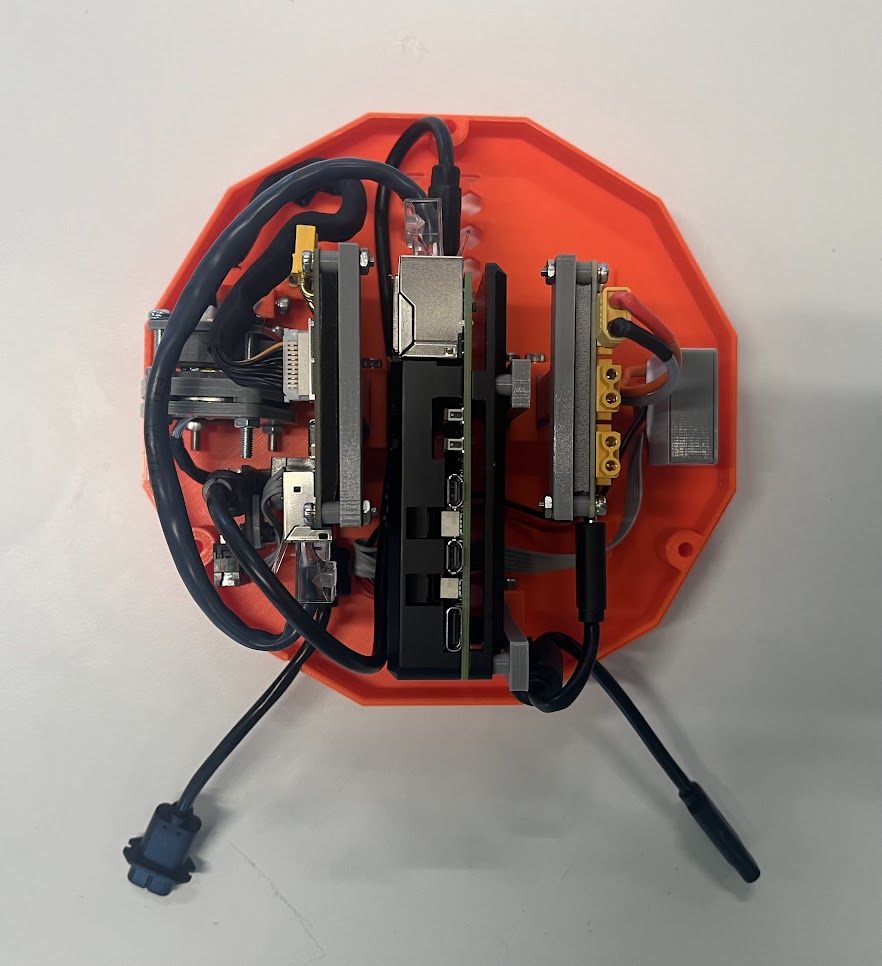}
    \includegraphics[height=4cm]{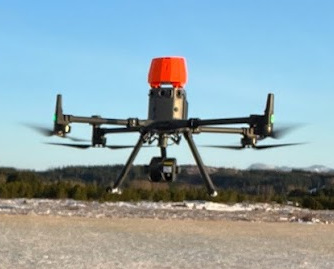}
    \caption{
        Payload case and component layout (left) and payload in flight (right).
        The payload case contains the Raspberry Pi, E-port and SkyPort development boards,
        SkyPort V2 connector (disassembled to save space), and DC-DC converter for regulated
        power supply.
        It mounts modularly and can be added/removed as needed.
        This relatively small payload allows us
        to fully utilize the sophisticated drone, i.e.,
        to access the drone's sensors and cameras,
        control the drone and other payloads autonomously,
        and process/stream video to the controller in real time.
    }
    \label{figure:payload_case_component_layout}
\end{figure}

We mount all the components in a 3D-printed case that attaches at the payload mount
point on top of the drone and fits in the drone's case, meeting \textbf{Req.~\ref{requirement:dji_style_portability}}.
Figure~\ref{figure:payload_case_component_layout} shows the component layout
and in-flight visibility of the payload.
The E-port application can access the feeds from sensors (\textbf{Req.~\ref{requirement:onboard_sensors}}) and RGB and stereo cameras on the drone and from other payloads (\textbf{Req.~\ref{requirement:onboard_video_feeds}}),
and can control the behavior of the drone and of other payloads (\textbf{Req.~\ref{requirement:control_drone_and_payloads}}).
The primary difficulty for this application is the setup,
which is not clear from the documentation.
The SkyPort application is able to stream the desktop such that it can both display
the visualizations from the E-port application (\textbf{Req.\ref{requirement:process_transmit_video}}).
To keep the load on the Raspberry Pi's CPU at or below 10\% while providing
adequate video quality, we set the desktop resolution to 640x480 pixels and
transmit at 24 FPS.
We configure \lstinline{ffmpeg} to encode with the ``ultrafast'' preset
using a bitrate of 1.2 Mbps, and a ``GOP'' (``Group of Pictures'') of 2,
such that the streamer alternates between sending keyframes and I-frames,
which makes the video transmission smoother and more reliable,
though not as lightweight as it could be.
We do not send B-frames.
This results in a latency of about 300 milliseconds before events on the Pi are
displayed on the controller's screen.
This is most likely due almost solely to the slow software encoding on the Pi,
since the high-definition video streams from the other cameras have almost
no latency.

\begin{figure}
    \centering
    \includegraphics[width=0.8\columnwidth]{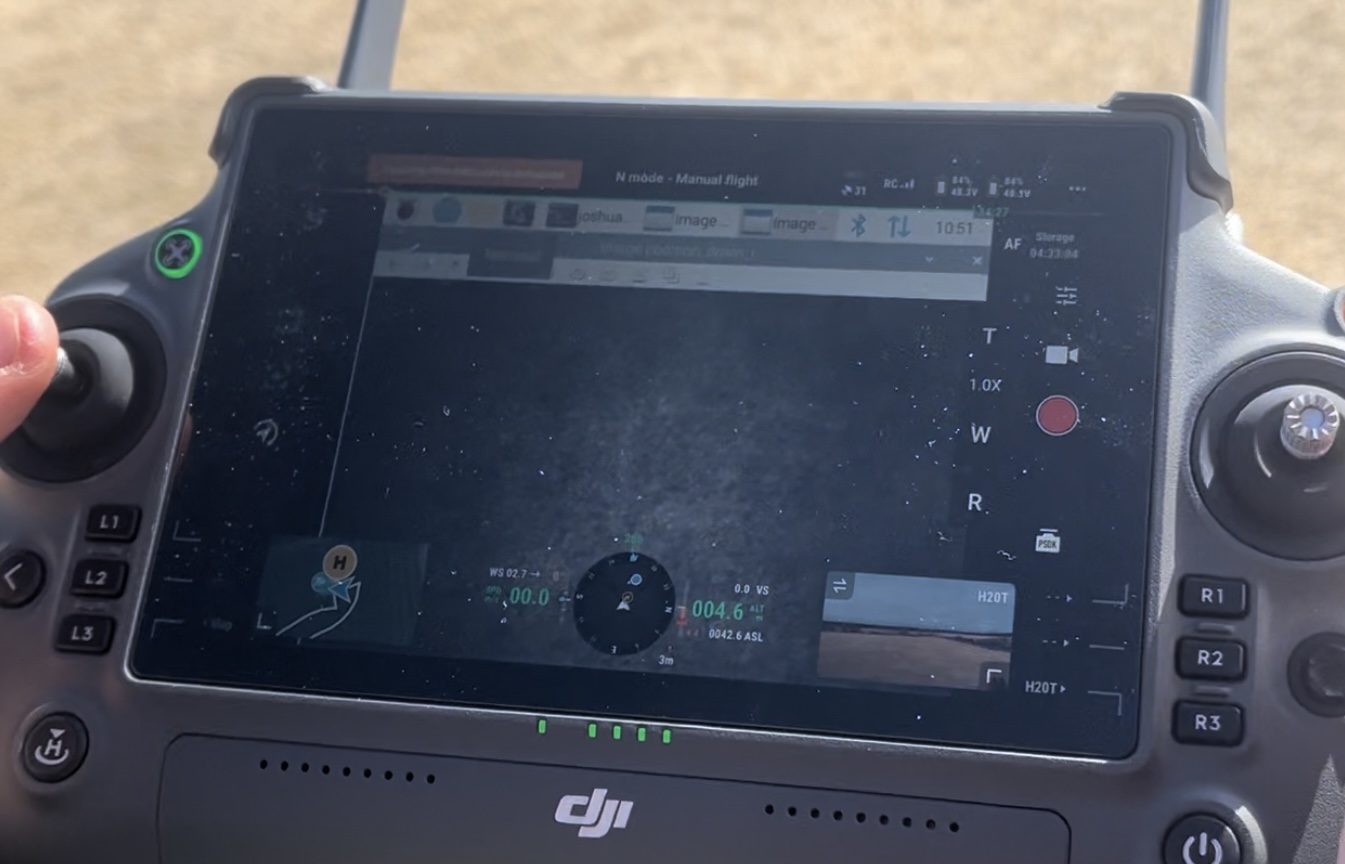}
    \caption
    {
        In-flight view of the downward stereo camera feed from the Pi's desktop.
        The image shows the Pi's taskbar on the top, with the video window below.
        The Pi is selected as a payload device,
        and we can switch between the Pi and the H20T (bottom right).
        Touching the screen at a particular spot causes a corresponding click on the Pi's screen.
    }
    \label{figure:rpi_payload_desktop}
\end{figure} 

\subsection{Quirks}

The lower-level functions of the DJI Payload SDK are a black box and
some of its behaviors can be unexpected at first.
The following is a list of such issues we have addressed.
\textbf{Q1:} The desktop streaming application has to start after the Pi has fully initialized;
we have found that reaching a minimum uptime of 180 seconds works reliably.
Otherwise, the encoding and streaming is unstable and causes crashes.
\textbf{Q2:} The SkyPort application needs to start before the E-port application.
During payload negotiation, the PSDK configures the high bandwidth communication
channel over the \lstinline{eth0} interface,
but also carries out some other, unknown behavior that breaks the E-port
application's high bandwidth data channel over the USB bulk devices.
We have similarly found difficulties in using USB-to-UART converters to handle the 
UART communication on both apps;
starting the second application crashes both converters,
even though they are addressed differently.
We have found that it is best to decouple the interfaces that the applications use
as much as possible,
so we use the Pi's onboard UART and the USB bulk devices for one,
and a USB-to-UART converter and Ethernet interface for the other.
\textbf{Q3:} While it is not mentioned in the documentation,
it is not possible to retrieve the stereo video streams over the E-port device 
when using a network device such as ethernet over USB~\cite{bulk_link_only_for_stereo}.
It is only possible using the USB bulk devices,
but the only tutorial for setting this up is for the Matrice 30,
which uses a single USB bulk device.
This led to our adaption in Section~\ref{section:eport_application}.
\textbf{Q4:} The SkyPort V2 connector, which is necessary for connecting to the drone's
PSDK port, is not listed as being compatible with the Matrice 350 on the product's
website, although it is compatible -- after a firmware update.
However, the firmware update must be carried out on a Matrice 300, not Matrice 350.
We conducted the update via the local DJI store, but it failed and bricked the SkyPort connector
on the first attempt, using the latest version of the DJI Assistant software.
It succeeded later on an earlier version of the software provided by DJI for this purpose.
Each link in this chain required communication with DJI via its developer forum,
with a latency of 1-2 business days.
\textbf{Q5:} Everything from DJI's drones to their batteries
and charge cases have microprocessors and communicate to each other.
This complexity introduces unexpected difficulties, e.g.,
repeated runtime errors when testing code to transmit desktop video
to the controller propagated through the drone system deeply enough that the
batteries reported communication errors with the drone, which were solved through rebooting.

\section{Conclusion}
We have demonstrated a method for integrating custom computational payloads into the DJI
Matrice 350 using a Raspberry Pi 5.
The system can access the drone's sensor and video feeds, control the drone's flight and the behavior
of other payloads, process video and stream it to the controller in real time.
It can also provide basic interaction with the Raspberry Pi via mouse clicks on the controller's screen.
We have also provided the scripts to configure such a system, so that such functionality
may be within reach for researchers from other domains.
The payload is cheap and can be easily adapted to deploy other sensors supported by the 
Raspberry Pi 5.

\bibliographystyle{IEEEtran}
\bibliography{references}

\end{document}